\documentclass[final,12pt]{arxiv} 

\title[Geospatial Disparities: A Case Study on Real Estate Prices in Paris]{Geospatial Disparities: A Case Study on Real Estate Prices in Paris}
\usepackage{times}
\usepackage{tikz}
\usepackage{multirow}
\usepackage{dsfont}

\newtheorem{theo}{Theorem}[section]
\newtheorem{rem}[theo]{Remark}

\usepackage{algorithmic}
\usepackage{algorithm}

\usepackage{colortbl}

\usepackage{algorithm}
\usepackage{algorithmic}
\usepackage{colortbl}
\usepackage{booktabs}
\usepackage{float}

\usepackage{comment}

\newcommand{\argmax}[1]{\underset{#1}{\operatorname{arg}\!\operatorname{max}}\;}

\clearauthor{%
\Name{Agathe Fernandes Machado} \Email{fernandes\_machado.agathe@courrier.uqam.ca}\\
   \addr Université du Québec à Montréal
   \AND
 \Name{Fran\c{c}ois Hu} \Email{francois.hu@umontreal.ca}\\
 \addr  Université de Montréal
 \AND
  \Name{Philipp Ratz} \Email{ratz.philipp@courrier.uqam.ca}\\
  \addr Université du Québec à Montréal
   \AND
   \Name{Ewen Gallic} \Email{ewen.gallic@univ-amu.fr}\\
   \addr Aix Marseille Univ, AMSE, CNRS, France
   \AND
 \Name{Arthur Charpentier} \Email{charpentier.arthur@uqam.ca}\\
 \addr Université du Québec à Montréal%
}

\begin{document}

\maketitle

\begin{abstract}%
Driven by an increasing prevalence of trackers, ever more IoT sensors, and the declining cost of computing power, geospatial information has come to play a pivotal role in contemporary predictive models. While enhancing prognostic performance, geospatial data also has the potential to perpetuate many historical socio-economic patterns, raising concerns about a resurgence of biases and exclusionary practices, with their disproportionate impacts on society. Addressing this, our paper emphasizes the crucial need to identify and rectify such biases and calibration errors in predictive models, particularly as algorithms become more intricate and less interpretable. The increasing granularity of geospatial information further introduces ethical concerns, as choosing different geographical scales may exacerbate disparities akin to redlining and exclusionary zoning. To address these issues, we propose a toolkit for identifying and mitigating biases arising from geospatial data. Extending classical fairness definitions, we incorporate an ordinal regression case with spatial attributes, deviating from the binary classification focus. This extension allows us to gauge disparities stemming from data aggregation levels and advocates for a less interfering correction approach. Illustrating our methodology using a Parisian real estate dataset, we showcase practical applications and scrutinize the implications of choosing geographical aggregation levels for fairness and calibration measures.
\end{abstract}



\section{Introduction}



Predictive models are now ubiquitous, churning through huge amounts of data collected by an ever-increasing number of different sources, and in turn, provide more granular predictions. Particularly, geospatial data, surging in availability and granularity owing to remote sensing and IoT devices, has gained popularity within the Machine Learning (ML) community. At the same time, the evaluation of ML models no longer depend solely on performance metrics such as accuracy but currently also includes considerations of fairness and calibration within the predictions. A natural question that then arises from these observations is how the increased presence of geospatial data influences understanding of our assessment of model calibration and the measurement of fairness in predictions. 

An area where geospatial data has long been present in predictive models is real estate pricing. However, real estate and geospatial locations have a long history of discrimination, spanning from denying services, referred to as redlining \citep{nier1998perpetuation}, over exclusionary zoning, which attempts to limit economic and racial diversity in the first place, to gentrification. Although such practices are often outlawed in modern societies, the recent surge in the usage of algorithms has renewed concerns that their predictions could perpetuate biases present in current learned data. 

The field of Algorithmic Fairness has developed to understand and measure how specific demographic groups may face disparate impacts owing to biases with algorithms in a variety of settings \citep{agarwal2018reductions, Chzhen_Denis_Hebiri_Oneto_Pontil20Wasser,denis2021fairness,hu2023fairness}. However, measuring disparities within communities defined by spatial proximity remains challenging, partly because spatial proximity is often not clearly defined. Additionally, fairness itself is difficult to define given spatial locations, as location can often proxy for a range of variables, which might be legitimately used in a model on one side or be the source of undue discrimination on the other side.

Closely linked to fairness considerations are issues that emerge due to the (mis-)calibration of a model, which results in systematic deviations of predictions from true probabilities and compromises overall reliability. Again, the effects of miscalibration can be distributed differently depending on geolocation, which may result in new sources of unfairness and biases. Hence, when evaluating predictive power, a meticulous examination of the disparities between predicted and observed values is paramount to determine whether the model exhibits underconfidence or overconfidence \citep{brahmbhatt2023towards}, especially with respect to given subgroups. The issue of a non-calibrated model is further concerning for a variety of prediction tasks, as decision makers might rely heavily on the initial valuation of predictive models for their final decisions, leading to the anchoring effect \citep{Tversky1974Science} or misleading interpretation of the given initial valuation.

As modern algorithms become more complex and less interpretable, regulators have started requiring higher standards of models and their associated outputs. Notable examples include the GDPR (2018) and the upcoming EU AI Act (2024) in Europe with heightened scrutiny of the justified use of geospatial data. In light of the aforementioned concerns about both calibration and fairness, one of our core objectives is to comprehensively understand what actually constitutes ``spatial biases'', whether stemming from intentional choices or unintentional factors, as outlined in the Algorithmic Fairness or Bias literature. We aim to propose an effective framework to mitigate these effects and ensure the ethical deployment of complex algorithms.

To stay close to a realistic scenario involving a decision maker or regulator, wherein both the training data and the learning algorithm are not directly accessible, we conduct our study purely on geospatial indicators, model predictions, and collected labels in an \textit{ex-post} study. Using a real-world dataset, we present a process to evaluate when and how disparities correlate on a spatial level and examine what this implies for both calibration and fairness. For cases in which biases can be detected, we propose a simple \textit{ex-post} correction of the model's predicted values to ensure compliance with both calibration and fairness while minimizing the overall effect on the predictive quality. A fundamental issue for this analysis is the choice of the aggregation level for geospatial effects. On one extreme, the most granular representation could be used; however, given the limited availability of data, this would most likely lead to estimates with large variances. On the other extreme, aggregating data at the highest level would mask most of the insights that a more granular representation could provide \citep{holtgen2023richness}. Any inquiry is further complicated by the absence of consistently defined units of aggregation in most datasets, where regions can have different sizes or densities. 

\begin{figure}[htbp]
    \vspace{.1in}
    \centering
    \includegraphics[width=.7\textwidth]{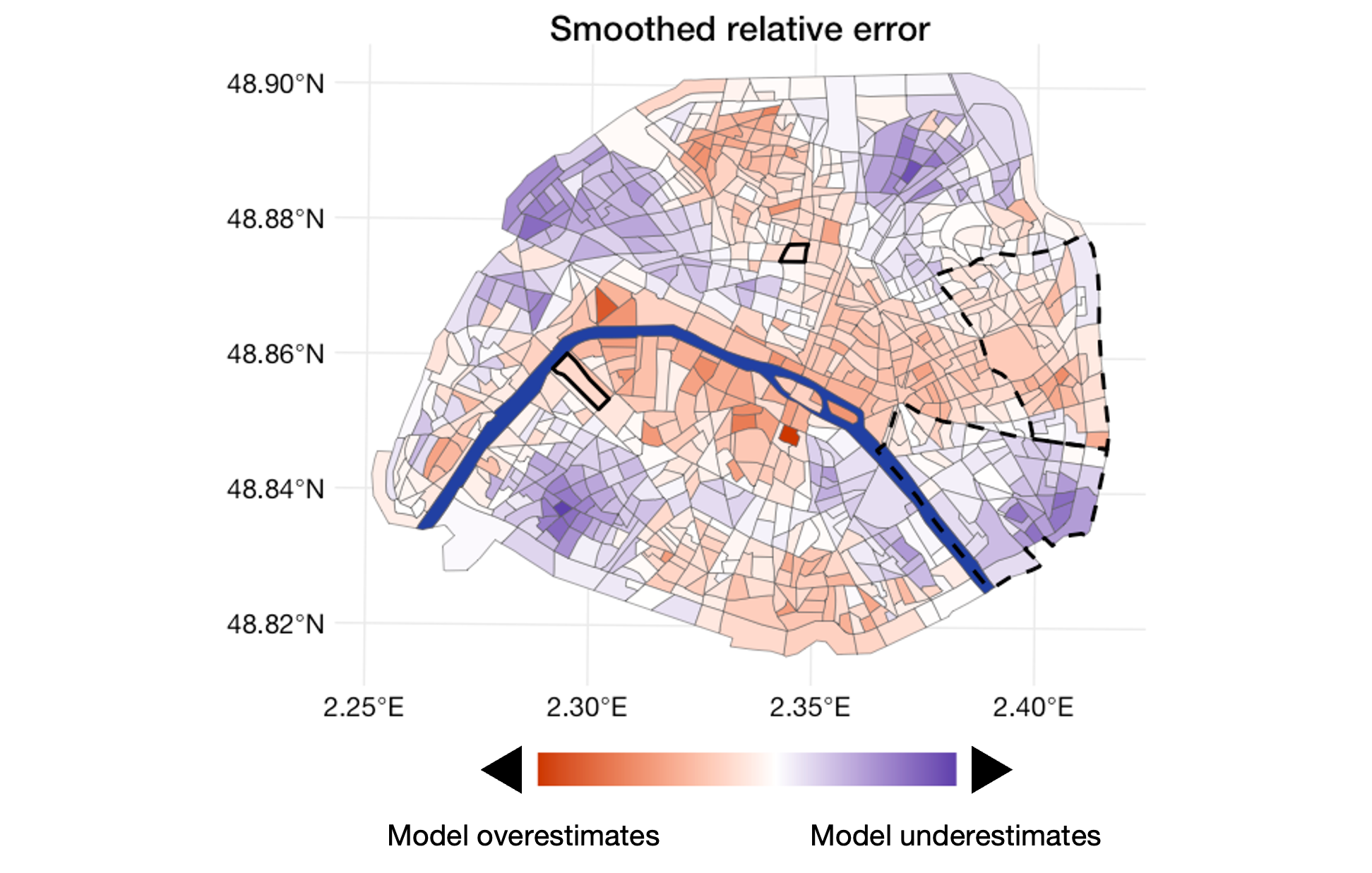}
    \caption{
    Relative estimation error per $m^2$ in different sub-regions. The values are smoothed across spatial neighbors to emphasize the spatial correlation as outlined in Section~\ref{sec:smoothing}.}
    \label{fig:smoothed_diffs}
\end{figure}

Considering the absence of a universally accepted level of analysis, we set out to evaluate how the choice of this level affects our results. As a motivational example, consider Figure~\ref{fig:smoothed_diffs}, which depicts the relative error of a model in estimating square meter prices. Although the model seems to be extremely well-calibrated with small overall errors, there is a spatial correlation among the residuals. However, evaluating the underlying causes of this situation cannot easily be characterized by administrative units, demarcated by dashed lines in the graph, and commonly used by decision makers.

Finally, as mentioned above, geolocation paired with housing has a long history of undesirable disparities, which makes the use of location in predictive models prone to perpetuating biases. Hence, identifying geospatial biases should be one of the priorities for policymakers. Although the socioeconomic status of neighborhoods can and does change over time, this change is often slow, as documented by \cite{rosenthal2015change}. For instance, \cite{heblich2021east} showed that pollution levels from the 1800s could still predict the share of low-skilled workers in a neighborhood, even after the removal of industrial sites responsible for pollution. If such effects can persist in time, using models trained to replicate past events might only reinforce existing disparities. 






\subsection{Main contributions}

Motivated by the aforementioned ethical concerns, including income segregation, and with the objective of deepening our understanding of geospatial disparities, their potential negative impacts, and viable mitigation strategies, we conduct a case study focusing on the Parisian real estate market. Our analysis involves comparing the property estimates provided by a real estate agency with actual observed sale prices. Real estate pricing methodologies are extensively studied and pricing usually depends on a large number of factors, both directly observable and factors that need to be proxied for. At the same time, the real estate market has been extensively studied from the perspective of segregation and other social aspects \citep{chambers_1992_J_Urban_Econ,Kieli_1996_JHE,Myers_2004_J_Urban_Econ,Bayer_2017_J_Urban_Econ}. This makes the field an ideal testing ground for the development of algorithms that assess geographical disparities in a more general framework. 

Although centered on a specific case study, our approach remains agnostic to the type of predictive models or the variable that is predicted. This broadens the scope of our analysis to include a more general setting. For example, one might examine predictions for the required number of child daycare spots compared to actual application numbers, or similarly, anticipated passenger volume in contrast to the total journeys taken. In each scenario, disparities in calibration and fairness across regions could lead to variations in recommended policy actions.


More specifically to our case study, we are interested in the error rates of the model. The average observed price difference is €203 per square meter (or €1179 in absolute terms) relative to a median sales price of €10,388 per square meter. Although this error is extremely small in a global context, Figure~\ref{fig:smoothed_diffs} shows that these errors are not necessarily uniformly distributed, suggesting a potential inclination of the predictive model to overestimate sale prices in some neighborhoods and underestimate them in others, even for properties with similar attributes. Given the anchoring effect of an initial valuation, this might lead to undesirable outcomes. Although estimated and realized sale prices are expressed on a continuous scale, we opted for the discretization of values to facilitate the analysis. This approach situates us in an ordinal regression context, enabling the calculation of fairness metrics such as Equalized Odds. The ordinal regression context does not preclude the assessment of the model calibration.

In summary, we find that even if a model is globally well-calibrated, there can be significant differences in fairness metrics in distinct sub-regions. Furthermore, these differences depend on the size and shape of the area under consideration, which pose issues when not all decisions are based on the same aggregation level. To counteract this, we extend previous research on post-processing methods to achieve fairness over a variety of regional groupings. This allows us to change the scores that can be used for policy decisions regardless of the level of analysis. To help researchers in the future, we propose a standard toolbox to measure and mitigate biases based on geospatial datasets.






Specifically, the contributions of the present article can be summarized as follows:
\begin{enumerate}
    \item We conduct an examination of model errors using post-processing approaches, revealing nuances in predictive biases.

    \item We contribute by developing a dedicated framework highlighting geospatial disparities induced by predictive models, aiming to improve algorithmic fairness in geographic contexts and improve the overall understanding of the underlying dynamics.

    \item The introduction of a practical case study using Parisian real estate data empowers decision makers to scrutinize disparities in proposals, fostering informed decision-making, and ensuring equitable assessment of geographic considerations.
\end{enumerate}

We begin by providing an overview of the geographic disparity problem and introducing the associated notation in Section~\ref{sec:problem_statement}.
Subsequently, in Section~\ref{sec:detecting-geo-disparities}, we conduct numerical experiments using Parisian real estate data to showcase the intricacies of the issue and to derive associated insights. Within our comprehensive bias mitigation framework, we illustrate, by an example in Section~\ref{sec:detect-and-mitig}, how various post-processing techniques can be applied within our contextual framework.



\subsection{Related work}

Much of this work aligns with the literature on biases and fairness with geographical considerations. The notion of ``(geo)spatial bias'' resonates with the previously explored notion of ``spatial equity'' (or ``spatial fairness'') in \cite{hay1995concepts}, which emphasizes the fair distribution of resources and opportunities across various geographical areas. This concern has attracted the attention of researchers from different domains, spanning from public health, where the emphasis lies on ensuring an equitable allocation of medical and health resources to the public \citep{bigman2000spatial, mattei2018fairness}, to urban greenery, where endeavors are directed towards developing urban parks that transcend socioeconomic and ethnic boundaries \citep{comber2008using, tan2017effects, yang2020understanding}. The goal of understanding \citep{ratz2023fairness, hu2023sequentially} (or mitigating \citep{chzhen2022minimax, hu2023fairness, charpentier2023mitigating}) these biases is to scrutinize biased scores (or transform them into equitable ones). Following a simple privacy-preserving framework, we study spatial disparities with limited information about descriptive features and the learning process of the predictive model. However, we retain data on the geographical location, true score, and model score, as detailed in the real estate case study in Section~\ref{sec:detecting-geo-disparities}. This justification supports the application of post-processing techniques that can be implemented in addition to any predictive model.






\section{Problem statement}\label{sec:problem_statement}

In predictive modeling, standard tasks include regression for predicting real-valued outputs and binary classification for categorization into two classes. In the regression task, predicting real values poses challenges for label-conditional modeling or measurement because of the nature of continuous random variables (no mass), which hinders the assessment of the \emph{Equalized Odds} notion of fairness defined in the next section. To simplify theoretical considerations and enable the comparison of classes within a distribution, one effective approach is to discretize the task into a set of bins while preserving the order of the bins. 
This practice, often used in decision-making, is known as the \emph{ordinal regression} case (see~\cite{gutierrez2015ordinal}).
Our objective is to predict outcomes within an ordered set $\mathcal{Y} := [K] = \{1, \ldots, K\}$---also known as a form of multi-class classification (see \citealt{tewari2007consistency, kolo2011binary}). More specifically, we aim to understand how predictors impact different response levels. We denote $\mathcal{X}\subset \mathbb{R}^d$ the feature space and we let $\mathcal{H}$ be the set of all predictors of the form $h:\mathcal{X}\to \mathcal{Y}$. 

Given our assumption that both the predictive model \(h \in \mathcal{H}\) and the data \(\boldsymbol{X} \in \mathcal{X}\) are unknown (at least partially), the choice of ordinal regression allows for a more robust approach that is especially effective against skewed distributions with well-positioned cuts for generating bins. 

Additionally, we define a sensitive attribute as a characteristic that is considered sensitive because of its potential to introduce bias or discrimination in decision-making processes. This sensitive attribute, denoted \(A \in \mathcal{A}\) with \(\mathcal{A} := [M] = \{1, \ldots, M\}\) is herein defined as a discrete group representing specific geographic locations, also known as \emph{Geospatial Data}. 
The sensitive attribute $A$ represents a geographical segmentation for which obtaining fair predictions is desirable. This choice of segmentation may be driven by the nature of the data. Certain geographic indicators may be provided based on one partitioning, for example, according to political zone delineations, whereas others may be provided based on a different partitioning, for example, according to administrative delineations. When creating the dataset used for the predictive task, the modeler must make choices to group values on a common spatial scale. Additionally, to comply with data protection and anonymity rules, the values provided on a certain geographical scale may result from a different spatial aggregation from one geographic area to another. When a geographic area lacks a sufficient number of observations, the organization responsible for their dissemination must resort to spatial aggregation to ensure anonymity. Whether the aggregation is due to common scaling or compliance with legal rules, the choices made to define geographic segmentation may not be inherently neutral. This naturally creates a conundrum for any analysis, because the aggregation level of the data can either hide or emphasize certain aspects inherent in the data. 
In this context, our aim is to investigate whether a well-calibrated model might exhibit bias based on geolocations. To achieve this, we will introduce metrics designed to assess both model \emph{calibration} and model \emph{unfairness}.


\begin{rem}[The term ``fairness'']
    In lieu of the term ``fairness'', using a more neutral expression such as ``unbiased'' or equivalent expressions for ``unfairness''would be more apt to use in our discussion. The intent behind using (un-)fairness is not to suggest a discriminatory usage of the data \emph{per se} but rather to emphasize the inequity in under-evaluation. It's crucial to note that the term ``unfairness'' is employed here without a negative connotation, as a not fair outcome might be positive for some. The decision of using the term ``unfairness'' is made to maintain consistency with the established language in the algorithmic fairness literature, where this term is commonly used to denote disparities, despite the absence of an inherently negative implication in our specific context.
\end{rem}



Throughout this article, we represent real-valued tasks as $Z$ or $\hat{Z}$ and multi-class tasks as $Y$ or $\hat{Y}$.

\subsection{Background on Model Calibration}

\label{subsec:calibration}

In traditional regression, a model output denoted as $\hat{Z}$ is considered well-calibrated for $Z$ when \citep{calib2019reg}:
$$
\mathbb{E}\left[ Z \mid \hat{Z} \right] = \hat{Z}\enspace.
$$
Let $(\boldsymbol{X}, Y)$ be a random tuple with distribution $\mathbb{P}$. In the context of our study, where we focus on ordinal regression denoted as $\mathcal{Y} = [K]$, where $K>0$ is a positive integer, a model $h\in\mathcal{H}$ is deemed well-calibrated in predicting the distribution of confident scores $\hat{\boldsymbol{P}}$ \citep{widmann2019calibration} if:
$$
\mathbb{P}(Y = k | \,\hat{\boldsymbol{P}} = \hat{\boldsymbol{p}}) = \hat{p}_k,\quad \text{ for all }k\in[K]\enspace,
$$
where $\hat{\boldsymbol{p}} = (\hat{p}_1, \ldots, \hat{p}_K)$ and $\hat{p}_k$ represents the (confidence) score or probability estimate for class $k$ under the model $h$. This calibration concept extends to group-wise calibration when considering a sensitive attribute \citep{widmann2019calibration, wang2020towards}. In this scenario, a \textbf{strongly} calibrated model is achieved when:
$$
\mathbb{P}(Y = k | \,\hat{\boldsymbol{P}} = \hat{\boldsymbol{p}}, \,A = a) = \hat{p}_k,\quad \text{ for all }k\in[K] \text{, and for all }a\in A\enspace.
$$
In other words, the model is strongly calibrated if the conditional probability of the true class being $k$ matches the predicted probability for all classes $k$, for all geographical regions $a$.
This definition can be undermined to acquire a \textbf{weakly} calibrated model:
$$
\mathbb{P}(Y = \arg\max_{k} \hat{P}_k \, | \,\max_{k}\hat{P}_k = \hat{p}_k, \,A = a) = \hat{p}_k,\quad \text{ for all }a\in A\enspace.
$$
In simpler terms, weak calibration means that the highest predicted probability is used as a threshold and the model is considered calibrated if the true class matches the highest probability.

Furthermore, we introduce the group-wise measure of calibration error, known in the calibration literature as the Expected Calibration Error (ECE)~\citep{naeini2015obtaining}. We specifically measure group-wise calibration in its weak form, denoted as $\mathcal{U}_{ECE}$, as presented below. Note that, in practical scenarios, the distribution $\mathbb{P}$ is not known. Instead, given empirical data $(\boldsymbol{X}_i, Y_i)_{i=1, \ldots, n}$ as $n$ \emph{i.i.d.} copies of $(\boldsymbol{X}, Y)$, we work with the empirical distribution $\hat{\mathbb{P}}_{(\boldsymbol{X}_i, Y_i)_i}$ (or $\hat{\mathbb{P}}$ for ease of reading), defined as
$$
\hat{\mathbb{P}}:= \frac{1}{n} \sum_{i=1}^n \delta_{(\boldsymbol{X}_i, Y_i)}\enspace.
$$
Let the interval $[0,1]$ be partitioned into $B$ bins based on quantiles of $\max_k \hat{p}_k$ values, where each bin $b\in[B]$ is associated with the set $\mathcal{I}_b$ containing the indices of instances within that bin. This partitioning is used in the subsequent definition of the ECE, which is applicable to the multi-class classification framework and thereby extends to ordinal regression.
\begin{definition}[Model calibration in multi-class classification]
To quantify the ECE, we introduce the accuracy and confidence measures within each bin $b\in[B]$:
\begin{equation*}
    \text{\textit{(accuracy)}}\ \   \textrm{acc}(\mathcal{I}_b) = \hat{\mathbb{P}}_{(\boldsymbol{X}_i, Y_i)_{i\in\mathcal{I}_b}}(Y=\hat{Y}) \quad ; \quad  \text{\textit{(confidence)}}\ \ \textrm{conf}(\mathcal{I}_b) = \hat{\mathbb{E}}_{(\boldsymbol{X}_i, Y_i)_{i\in\mathcal{I}_b}} (\max_k \hat{p}_k)\enspace.
\end{equation*}
The calibration error measure for a model $h\in\mathcal{H}$ is then defined as
$$
\mathcal{U}_{ECE}(h) := \frac{1}{n} \sum_{b\in[B]} |\mathcal{I}_b| \cdot \left|\textrm{acc}(\mathcal{I}_b) - \textrm{conf}(\mathcal{I}_b)\right|
\enspace,
$$
and model $h$ is considered (group-wise) well-calibrated i.f.f. $\mathcal{U}_{ECE}(h) = 0$.
\end{definition}

The above definition permits the measurement of calibration within a multi-class classification framework, achieved by evaluating the model's confidence in predicting a specific class against the actual frequency of that class in the observed events. This calibration metric can be calculated within any subregion that includes a sufficient number of datapoints for binning, which is crucial for understanding the far off predictions in a more localized framework. In real estate valuation, where the initial model price acts as an anchor for future decisions, achieving good calibration is therefore essential. Although some error is to be expected, localized variations in error rates may lead to systematic over- or under-valuation in specific areas due to differences in the locally prevalent price category, even if the model is globally well-calibrated. To consider such disparities more specifically, we extend this study beyond calibration alone, with a specific focus on algorithmic fairness and underlying geospatial disparities.

\subsection{Background on Algorithmic Fairness}
\label{subsec:algofairgeo}





In the present article, we consider two types of fairness evaluation: \emph{Demographic Parity} \citep{calders2009building} (DP), which asks for independence of the predictive model from the sensitive attribute, and \emph{Equalized Odds} (EO) \citep{hardt2016equality}, which seeks independence conditional on all values of the label space. These definitions, classically considered in binary classification tasks, naturally extend to the multi-class classification framework, as demonstrated in \cite{alghamdi2022beyond} and \cite{denis2021fairness}.

\subsubsection{Demographic Parity} We let $\hat{Y}$ be the output of the predictive model $h\in\mathcal{H}$ defined on $\mathcal{X}$. From the algorithmic fairness literature, we define the (empirical) unfairness under DP as follows:
\begin{definition}[Fairness under Demographic Parity]
The unfairness under DP of a classifier $h$ is quantified by
    $$
\mathcal{U}_{DP}(h) := \max_{a\in\mathcal{A}, k\in[K]} \left|\, \hat{\mathbb{P}}(\hat{Y} = k | ±\, A = a) - \hat{\mathbb{P}}(\hat{Y}  = k)\, \right|\enspace.
$$
A model $h$ is called (empirically) exactly fair under DP i.f.f. $\mathcal{U}_{DP}(h) = 0$.
\end{definition}
Intuitively, DP serves as a widely adopted measure of unfairness that is applicable to various tasks, including regression and classification. This measure holds the advantage of being recognized in legal contexts and regulations. Nevertheless, in situations where the label $Y$ is assumed to be unbiased, there emerges a preference for a more nuanced measure of unfairness. Specifically, DP may hinder the realization of an ideal prediction scenario, such as granting loans precisely to those who are unlikely to default.

\subsubsection{Equalized Odds} We assume knowledge of the true and unbiased label $Y$. Another notion of fairness is EO, with its associated unfairness measure defined as follows:

\begin{definition}[Fairness under Equalized Odds]
The unfairness under EO of a classifier $h$ is quantified by
$$
\mathcal{U}_{EO}(h) := \max_{a\in\mathcal{A}, k, k'\in[K]} \left|\,\hat{\mathbb{P}}(\hat{Y} = k | \,Y \, = k', \,A = a) - \hat{\mathbb{P}}(\hat{Y} = k | \,Y = k'\,)\right|\enspace.
$$
A model $h$ is called (empirically) fair under EO i.f.f. $\mathcal{U}_{EO}(h) = 0$.
\end{definition}

Ultimately, considering a model $h\in\mathcal{H}$, our objective is to investigate biases related to the model calibration and unfairness defined above, specifically the measures $\left( \mathcal{U}_{ECE}(h), \mathcal{U}_{DP}(h) \right)$ or $\left( \mathcal{U}_{ECE}(h), \mathcal{U}_{EO}(h) \right)$. In the next section, we delve into the case study using Parisian real estate data, where we designate $\hat{Y}$ as the estimated price per $m^2$ and $Y$ as the sold price per $m^2$.


\begin{rem}[Achieving calibration and unfairness]
    Achieving group-wise calibration and EO simultaneously has been demonstrated to be impossible, except in highly constrained cases \citep{kleinberg2016inherent, pleiss2017fairness} or by relaxing the EO property to proportional equality, leading to the simultaneous optimization of both fairness and calibration \citep{brahmbhatt2023towards}. Similarly, incorporating notions of fairness through calibration is feasible when employing global calibration scores \citep{holtgen2023richness}. Despite this, the dependencies between calibration and fairness cannot be generalized, particularly with between-group calibration. Deviations from this measure may reveal unfairness in certain situations but not in others, depending on the specified definition of fairness \citep{loi2022calibration}. As a result, calibration and fairness have been studied independently, with fairness primarily assessed through the EO definition, among other factors.
\end{rem}

\section{Detecting Geographic disparities}\label{sec:detecting-geo-disparities} 

We present our main insights through a case study of Parisian real estate, in which we consider different levels of aggregation and highlight how both calibration and fairness-related metrics change across them. Our main goal is to study disparities localized in sub-regions, between predictions made, and labels obtained throughout a test period. In this context, we analyze how model error rates differ across spatial regions, and quantify them using fairness measures. Methods to mitigate some of these biases will be addressed in Section~\ref{sec:detect-and-mitig}.

\subsection{Data}

For our illustrations, we use data obtained from \emph{Meilleurs Agents}, a French Real Estate platform that produces data on the residential market and operates a free online automatic valuation model (AVM).\footnote{The source code can be found at \href{https://github.com/fer-agathe/parisian_real_estate/}{(https://github.com/fer-agathe/parisian\_real\_estate/)} (no data available).} We have access to both the estimated price of the underlying property and realized net sale price. We consider the realized sale price as the true underlying value $Z$ that we attempt to approximate with the model prediction $\hat{Z}$. Along with the realized and estimated prizes, we also have access to the approximate location and amount of square meters ($m^2$) of the property. In total, we obtained approximately 25,700 observations from the Paris Metropolitan Area, of which approximately 11,600 were located in the city of Paris, collected throughout 2019. We use the prices per $m^2$ to normalize the errors by property size. Further, we restrict our analysis to observations located within the city of Paris, as there appear to be significant differences in the per $m^2$ prices between the core city (Paris \emph{intra-muros}) and the surrounding areas. See Figure~\ref{fig:plot_price_diff} in the appendix for a visual representation. We also removed outliers with a price per square meter of over 20,000€ and observations from mostly commercial areas. In all, we then have access to 11,500 observations after these basic cleaning steps. 

Our Data contains geospatial information, aggregated at the \emph{IRIS} (\textit{Ilots Regroupés pour l'Information Statistique}) level, a statistical unit defined and published by the French National Institute of Statistics and Economic Studies. Each IRIS region represents a clearly defined area within France and many publicly available economic and societal indicators have been published on this level of granularity. The population within each unit generally consists of between 1,800 and 5,000 inhabitants, which live within a homogeneous living environment,\footnote{For more information on IRIS, refer to INSEE \href{https://www.insee.fr/en/metadonnees/definition/c1523}{(https://insee.fr/en/metadonnees/definition/c1523)}.} which makes this unit particularly suitable for our analysis. In total, our observations are divided across 878 iris regions.

\subsubsection{Defining Neighbors}\label{sec:neighbors}

\begin{figure}[htbp]
    \vspace{.1in}
    \centering
    \includegraphics[width=1\linewidth]{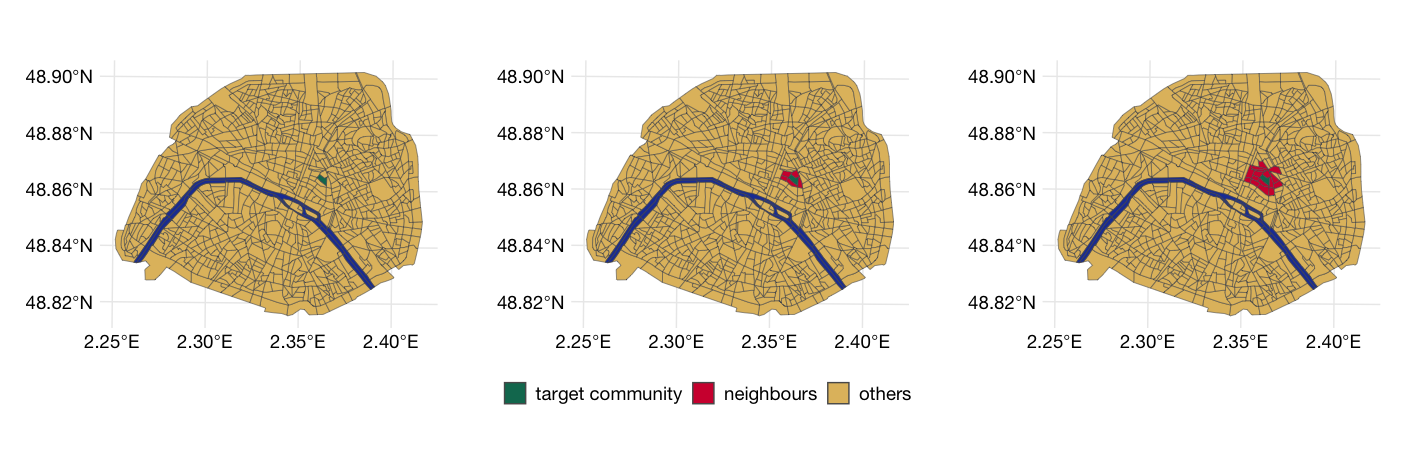}
    \caption{A sampled IRIS region within Paris (left pane) and its immediate adjacent neighbors (center pane) and the second level neighbors (right pane). The Seine River is depicted in blue whereas all other regions are depicted in yellow.}
    \label{fig:neighbourhood}
\end{figure}

The city of Paris is divided into 20 administrative units called \emph{arrondissements}, each grouping between 14 and 96 IRIS regions. It is important to note that an IRIS region cannot belong to multiple \emph{arrondissements}. Much information about real estate is typically aggregated at this level. However, for our analysis, this unit is too coarse, as there is still considerable heterogeneity within them as can be seen for example in Figure \ref{fig:smoothed_diffs}. Instead, we opt for a more flexible definition of spatial regions that takes advantage of the homogeneity within each IRIS region. As the size of the area of each IRIS differs considerably,\footnote{By a factor of almost up to 100.} simply defining higher levels by Euclidean distance from a given point might pose problems, as it could include many dense but heterogeneous regions on one end of the spectrum or only a few on the other. To take advantage of the differing sizes and homogeneity within the IRIS regions, we define a neighborhood graph. In its general form, a graph is an ordered pair $\mathcal{G}=(\mathcal{V}, \mathcal{E})$ of vertices and edges. In our application, each vertex represents an IRIS region, and each edge represents a neighboring relation. Here, a neighboring relation is defined as the presence of an intersection between two polygons, including their boundaries, defining an IRIS region. That is, edge $(i,j)$ is present in edge set $\mathcal{E}$ if the polygons of regions $i$ and $j$ intersect. The graph can then be represented as an $n\times n$ matrix $V$, called the adjacency matrix, where $V_{i,j}\in\{0,1\}$ with $V_{i,j}=1$ if $(i,j)\in \mathcal E$. That is, each entry in the adjacency matrix is equal to $1$ if two vertices are in a neighborhood relationship with each other. Representing neighborhood relations using an adjacency matrix has the advantage that non-intermediate connections can be easily obtained by successively multiplying the adjacency matrix by itself. That is, all nonzero elements of $V^2$ represent neighborhoods that are either immediately adjacent to a given region (the direct neighbors) or are adjacent to the direct neighbors (the neighbors of the direct neighbors). This process can be repeated $n$ times to obtain neighbors that can be reached within an $n$ length path. This provides a more natural way to define neighborhoods, as with each increasing path length, an entire homogeneous region is added to the higher-level aggregation. Figure~\ref{fig:neighbourhood} illustrates how the neighborhood set for a particular IRIS community can be calculated. 

\subsection{Neighborhood-Based Smoothing}\label{sec:smoothing}

To reveal spatial structures, as shown in Figure~\ref{fig:smoothed_diffs}, the raw data must be processed and filtered. One reason for this is that the observations are not uniformly distributed across IRIS regions. This naturally leads to different levels of confidence when summary statistics for each IRIS, such as the mean relative model error per $m^2$, are considered. The core idea of spatial smoothing is to use an average over a larger area, which should provide a more robust estimate. Applying a (weighted) mean function has the effect of a low-pass filter, which removes sharp edges between the IRIS regions, and hence produces an output that has a more pronounced spatial correlation, revealing the underlying structure. Kernel based methods are common within spatial analysis (see, \textit{e.g.}, \citealp{genebes2018spatial}). However, as discussed above, in our case, the data is already aggregated at the IRIS level, which restricts the application of spatial kernel-based smoothing whose bandwidth operates on Euclidean distances.

\begin{figure}[htbp]
    \vspace{.1in}
    \centering
    \includegraphics[width=\textwidth]{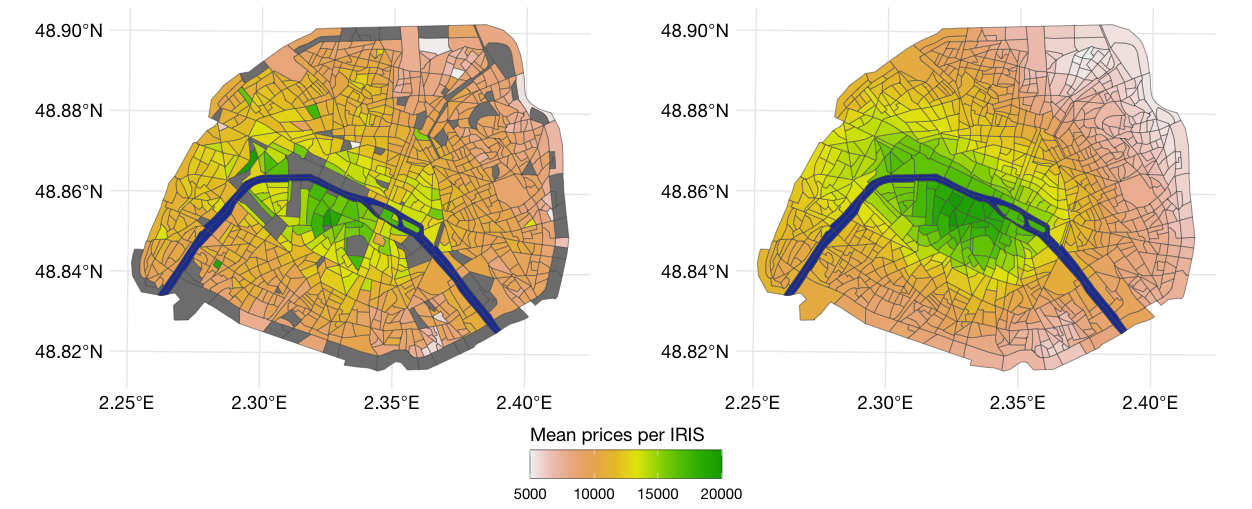}
    \caption{Smoothed square meter prices, corresponding roughly to the wealth level of the inhabitants. This serves as a motivation to analyze the predictions using quantiles in an ordinal regression framework, as it allows us to stratify the population according to socioeconomic status.}
    \label{fig:smoothed_prices_sqm}
\end{figure}

As an alternative, we use the constructed neighborhood graph and the path length between regions as the argument of a weight function, similar to kernel-based methods. That is, for a given variable observed at IRIS level $x$, the smoothed value for region $r_i$, denoted as $\lambda_\omega(x_i)$ can be written as:
\begin{align}
\lambda_{\omega}(x_i) = \frac{1}{\sum_{j=1}^{n} {\omega}(r_i,r_j)}\sum_{j=1}^n {\omega}(r_i,r_j) x_i \enspace.     
\end{align}
For example, let $d(r_i, r_j)$ be the path length between regions $r_i$ and $r_j$, then a simple way to define $\omega(r_i,r_j)$ is:
\begin{align}\label{eq:weights}
    \omega(r_i,r_j)=
        \begin{cases}
        \frac{1}{(1+d(r_i,r_j))^p},\quad & \text{if } d(i,j) \leq m \\
        0, & \text{otherwise} \enspace,
        \end{cases}
\end{align}
where $p$ and $m$ are hyperparameters to be selected, similar to the bandwidth operator. As an illustration, consider Figure~\ref{fig:smoothed_prices_sqm}, where we plot the observed prices per square meter directly. Many regions present sharp edges or missing data (in gray). Neighborhood-based smoothing can effectively interpolate these values and shows the typically observed pattern of prices, which are highest in central Paris on the south bank of the Seine River and lowest in the northeast of the city. Note that this map also roughly corresponds to the median income within the regions (see Figure~\ref{fig:smooth_med_inc} in the appendix for a visualization).

\begin{figure}[htbp]
    \vspace{.1in}
    \centering
    \includegraphics[width=\textwidth]{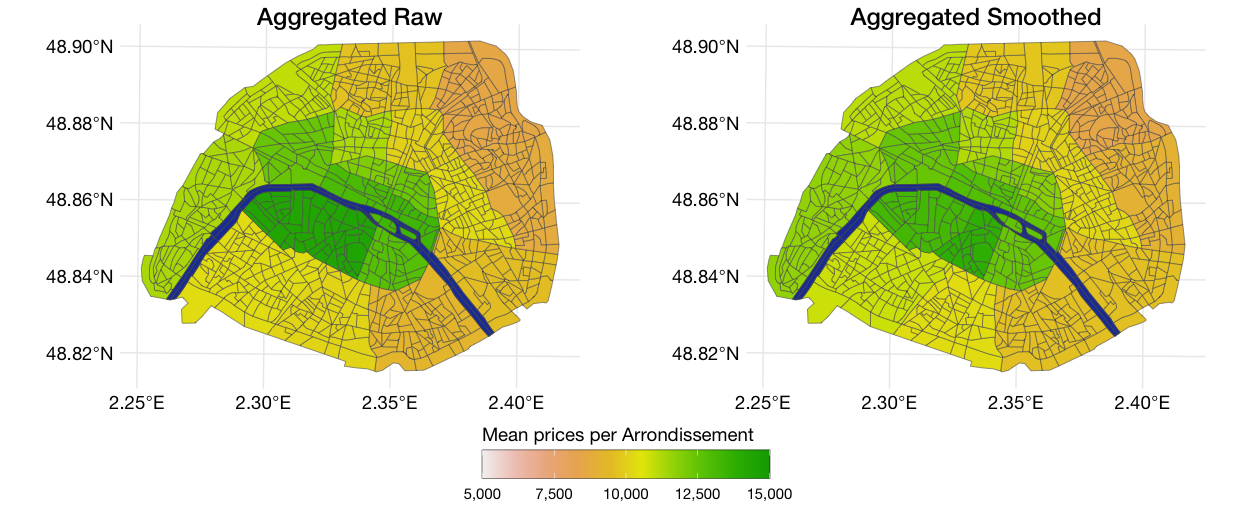}
    \caption{Re-Aggregated data, left pane, mean per \emph{arrondissement} when the raw, un-smoothed data is used to calculate the average price per square meter of real estate. Right pane, results when the neighbor-smoothed estimates are used. In general, there do not seem to be large differences between the methods, but the smoothed estimates allow easier and more robust inference.}
    \label{fig:re-aggregated}
\end{figure}

Although this smoothing allows us to perform a more robust and detailed analysis at the micro level, it does not seem to introduce larger distortions when re-aggregating at a higher level. For example, Figure~\ref{fig:re-aggregated} depicts the values obtained from both the raw data and smoothed values when calculating the mean square meter price per \emph{arrondissement}. Here, we used neighbors from a distance of up to five, that is, $m=5$ and $p=1$ in the weighting of Equation \eqref{eq:weights}. This allowed us to conduct a sensitivity analysis of the number of IRIS regions included when calculating either the fairness or calibration metrics. 

\subsection{Transforming continuous outcomes into ordinal regression predictions} \label{sec:conf_scores}


When dealing with a continuous outcome variable, such as predicting real estate prices, calibration can be assessed by segmenting the regression model predictions into bins or employing nearest neighbor methods on $\hat{Z}$. In this context, the initially continuous price per square meter in $\mathcal{Z}\subset \mathbb{R}$ is transformed into a multi-class variable in $\mathcal{Y}:=[K]$. The adaptability to customize the number of classes is illustrated in Figure~\ref{fig:calibration_bins_whole} in the Appendix.
For our analysis, we have selected a setup with $K=5$ classes, determined through quantile binning based on the observed range of price per square meter $Z$. We partition the variable set $\mathcal{Z}$ into classes using quantiles denoted as $\{q_0, \ldots, q_K\}$, where $q_k$ represents the $\frac{k}{K}$-th quantile of the data $(Y_i)_{i=1, \ldots, n}$. Each quantity within the interval $[q_k, q_{k-1}]$ is then assigned to class $k$, creating a meaningful segmentation of the data distribution in $\mathcal{Y}$.

In addition to transforming real values into classes, specific metrics, like the calibration measure emphasized by ECE in Section~\ref{subsec:calibration}, require the implementation of confidence scores. In line with our established notations, these scores, denoted as $\hat{p}_k$ for class $k$, are calculated using both the real-valued predictions $\hat{Z}$ and the transformed observations $Y$. 
To be more specific, these scores are determined based on the distances between each class midpoint and the estimated prices per square meter $\hat{Z}$. Subsequently, these distances are normalized by the length of each class interval. The \emph{softmax} function is then applied to these distances, providing confidence scores for all classes and observations.

\subsection{Results on Model Calibration}



\label{subsubsec:calibration}

\begin{figure}[htbp]
    \vspace{.1in}
    \centering
    \includegraphics[width=\textwidth]{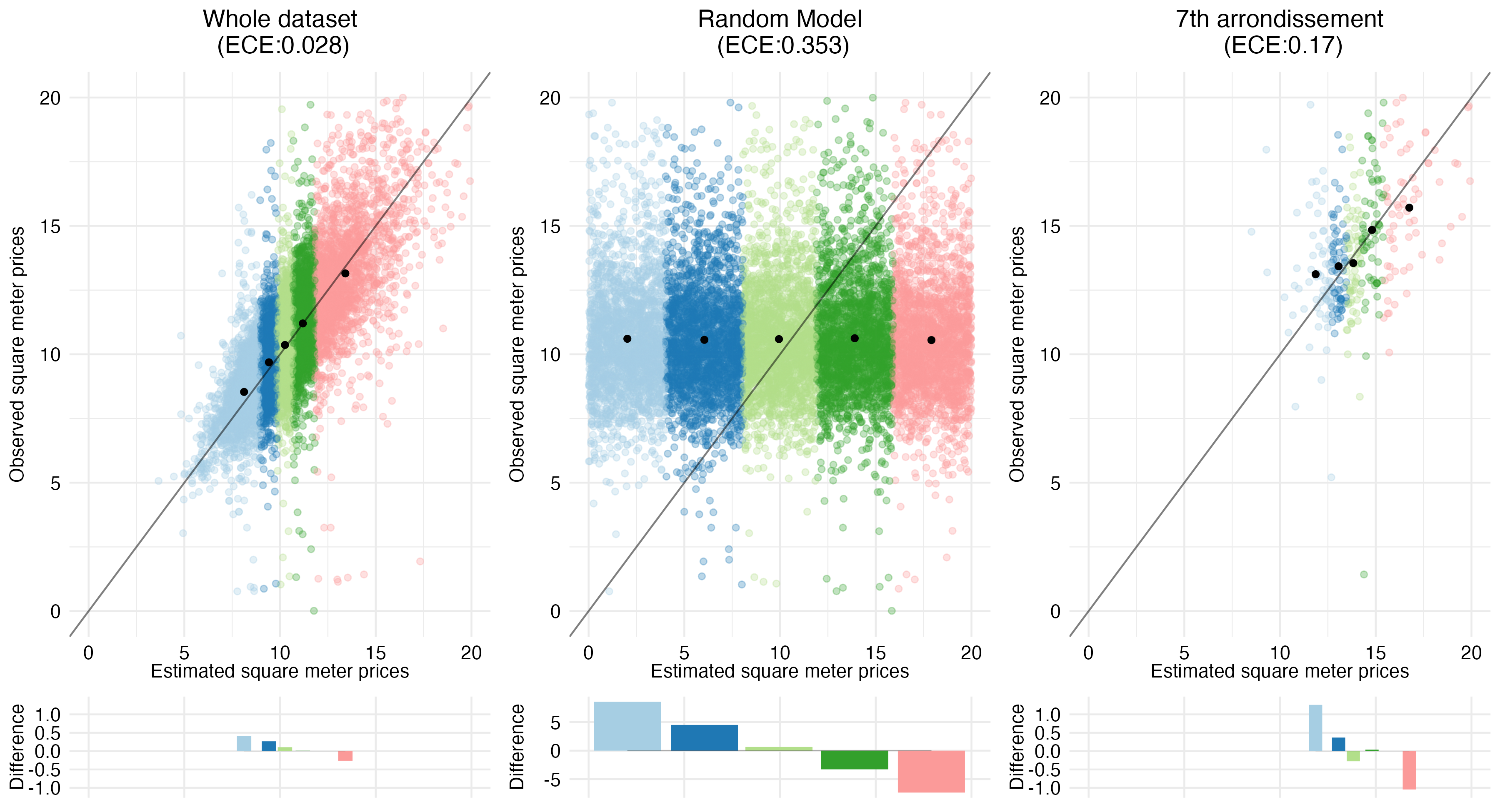}
    \caption{Calibration on the whole dataset  (left), on the observation from the 7th \emph{arrondissement} only (right) and on randomly drawn values (middle); bins defined using quantiles. Prices are in thousand Euros.}
    \label{fig:calibration_5-bins-comparison}
\end{figure}

The predictive model used to estimate real estate prices on the whole dataset appears to be well-calibrated, according to Figure~\ref{fig:calibration_5-bins-comparison}, when using quantile binning on the continuous estimated price. On average, the lowest square meter prices are slightly overvalued, whereas the highest are slightly undervalued. This suggests that the given model is generally well-calibrated. By focusing solely on the geographical information of each \emph{arrondissement} as a region of study, we observe slight miscalibration in certain areas, such as the 1st and 7th \emph{arrondissements} (we refer to Table~\ref{tab:ece_eo_arrond}), associated with a calibration error higher than that observed for all of Paris. Despite this, the 7th arrondissement, even with the highest ECE, demonstrates a level of calibration significantly superior to that of the random model.
This is illustrated in Figure~\ref{fig:calibration_5-bins-comparison}, where it can be observed that the predicted values for the 7th \emph{arrondissement} are closer to the observed values than for the random model, indicating a more calibrated model. 

Considering the well-calibrated nature of this model, we now aim to thoroughly assess its Algorithmic Fairness.

\subsection{Fairness results}

Note that the definition of $\mathcal{U}_{EO}(h)$ is defined over the set of \emph{feasible} quantiles.

\begin{figure}[htbp]
    \vspace{.1in}
    \centering
    \includegraphics[width=\textwidth]{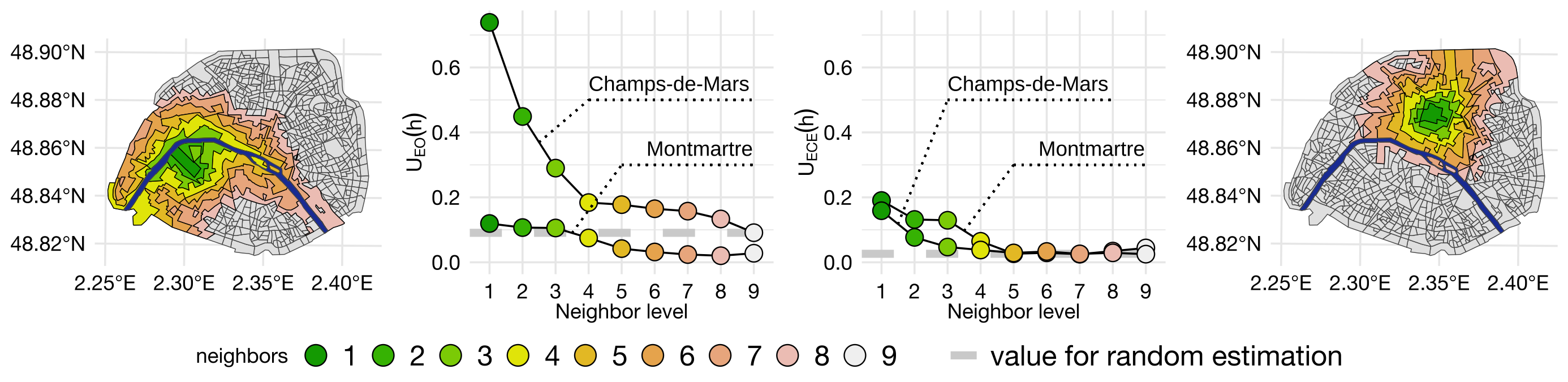}
    \caption{
    Equalized Odds Measure and Expected Calibration Error for different sizes of neighborhoods for two sample regions, \emph{Champs de Mars} (on the left) and \emph{Montmartre} (on the right).}
    \label{fig:eo_neighbours}
\end{figure}

\subsubsection{Overview by Administrative Units}

Each \emph{arrondissement} is characterized by unique levels of population density and wealth. Table~\ref{tab:ece_eo_arrond} presents the EO and ECE metrics computed for each \emph{arrondissement} of Paris. It is important to highlight that the fairness metric assesses EO by comparing a specific \emph{arrondissement} to the entirety of Paris, whereas ECE is calculated directly for each \emph{arrondissement}. An immediate correlation between the fairness and calibration metrics is not readily apparent, but certain \emph{arrondissements} share similarities in both measurements. For instance, the 7th \emph{arrondissement} exhibits poor performance in terms of both fairness and calibration errors, whereas the 14th and 15th \emph{arrondissements} rank among the best neighborhoods. Regarding calibration, this can be linked to both the number of observations in the dataset and, more generally, to population density, with the former \emph{arrondissements} having higher measurements, according to the French National Institute of Statistics and Economic Studies \footnote{For more information on the characteristics of \emph{arrondissements}, refer to INSEE \href{https://www.insee.fr/fr/statistiques/zones/1405599} {(https://www.insee.fr/fr/statistiques/zones/1405599)}.
}.

\subsubsection{Overview by Neighboring Classes}

Figure~\ref{fig:eo_neighbours} illustrates how the calibration (ECE) and fairness (EO) metrics expand as the level of spatial aggregation increases, starting from the two IRIS regions corresponding to Champs-de-Mars and Montmartre. The maps represent the areas considered at each level of grouping defined by the nearest neighbors approach (Section \ref{sec:neighbors}), starting with the first-level neighbors of Champs-de-Mars (left) and the same for Montmartre (right), ending with the neighbors of level nine. Observing the graphs depicting the ECE and EO for each of the two initial IRIS, it becomes apparent that as the level of aggregation increases, the metrics tend towards zero, which highlights the issue of gerrymandering. If a user wishes to calculate the calibration or unfairness of their model at a high level of aggregation, these will not be representative of unfairness measurements in sub-regions of the dataset, particularly for areas such as Champs-de-Mars. It should be noted that, based on the EO calculation, the random model (represented by the dashed line) is considered fair, as there is no distinction between subgroups due to its indiscriminate nature. For the various areas considered in Figure~\ref{fig:eo_neighbours}, a connection between fairness and model calibration emerges when examining the trends of the EO and ECE measurement curves.

\begin{table}[!htbp]
    \centering
    \caption{ECE on the whole dataset : 0.028}
	\label{tab:ece_eo_arrond}
    \renewcommand{\arraystretch}{1}
    \begin{tabular}{lllp{2em}lll}
	\hline\hline\\[-.5em]
	Arrond. & EO & ECE & &Arrond. & EO  & ECE \\ 
	\cmidrule(lr){1-3}\cmidrule(lr){5-7}
		1st & 0.301 & 0.188 && 11th & 0.108 & 0.136 \\ 
        2nd & 0.146 & 0.097 && 12th & 0.115 & 0.102 \\ 
        3rd & 0.319 & 0.126 && 13th & 0.177 & 0.052 \\ 
        4th & 0.469 & 0.116 && 14th & 0.070 & 0.089 \\ 
        5th & 0.280 & 0.131 && 15th & 0.072 & 0.072 \\ 
        6th & 0.637 & 0.126 && 16th & 0.107 & 0.055 \\ 
        7th & 0.625 & 0.170 && 17th & 0.085 & 0.096 \\ 
        8th & 0.128 & 0.157 && 18th & 0.145 & 0.029 \\ 
        9th & 0.117 & 0.120 && 19th & 0.396 & 0.078 \\ 
        10th & 0.114 & 0.132 && 20th & 0.242 & 0.088 \\
	\hline
    \end{tabular}
	\begin{minipage}{\textwidth}
	    \vspace{1ex}
	    \scriptsize\underline{Note:} Arrond. stands for \emph{arrondissement}, EO for Equalized Odds, and ECE for Expected Calibration Error %
	    \scriptsize\underline{Source:} Author(s) estimates.
	\end{minipage}
\end{table}

\section{Mitigating Geographic Disparities}
\label{sec:detect-and-mitig}

In this section, we delve into strategies to mitigate the biases mentioned earlier with a specific focus on addressing DP. Because our framework does not require knowledge of the predictive model or the data used during its learning process, post-processing mitigation approaches naturally emerge as the most suitable choices. In our example, we can leverage the constrained optimization approach proposed in \cite{denis2021fairness} to enforce the Demographic Parity notion of fairness. This post-processing method is tailored for multi-class classification tasks, aiming to achieve optimal fair classifiers with respect to misclassification risk under the DP constraint.

\subsection{Mitigation algorithm for Demographic Parity} \label{subsec:mitigdp}

Recall that $\hat{\boldsymbol{p}}$ denotes the confidence scores, and assuming the associated predictive model is fairness-aware, with $\hat{\boldsymbol{p}}(\boldsymbol{x}, a)$ defined on $\mathcal{X}\times \mathcal{A}$. In this section, we specifically consider $\mathcal{A} = \{1, 2\}$ as a binary set, where `1' denotes the 12th \emph{arrondissement}, and `2' represents the rest of Paris.
As per \cite{denis2021fairness}, one can readily recalibrate the outcome for exact fairness (for results in approximate fairness, we encourage readers to refer to the paper). Consequently, the new fair scores, denoted as $\hat{\boldsymbol{p}}^{(\text{fair})} = (\hat{p}^{(\text{fair})}_1, \ldots, \hat{p}^{(\text{fair})}_K)$ are then determined by
\begin{equation}
\label{eq:eqPlugIn}
\hat{p}^{(\text{fair})}_k(\boldsymbol{x}, a) = \hat{\mathbb{P}}(A=a)\cdot (\hat{p}_{k}(\boldsymbol{x}, a) - a \cdot\hat{\lambda}_k)\, ,\;\; \mbox{for all }(\boldsymbol{x}, a)\in \mathcal{X}\times {\mathcal{A}},
\end{equation}
where this technique is dedicated to enforcing DP-fairness of the overall classification rule by simply shifting the estimated conditional probabilities in an optimal manner, as dictated by the calibrated parameters $\hat{\boldsymbol{\lambda}} = (\hat{\lambda}_1, \ldots, \hat{\lambda}_K)$ where refer to the paper for more details about its optimization.
Given these new scores, the associated optimal fair predicted class is simply
$\argmax{k\in [K]} p^{(\text{fair})}_{k}$.

\subsection{Real Estate Case Study}

\begin{figure}[htbp]
    \vspace{.1in}
    \centering
    \includegraphics[width=\textwidth]{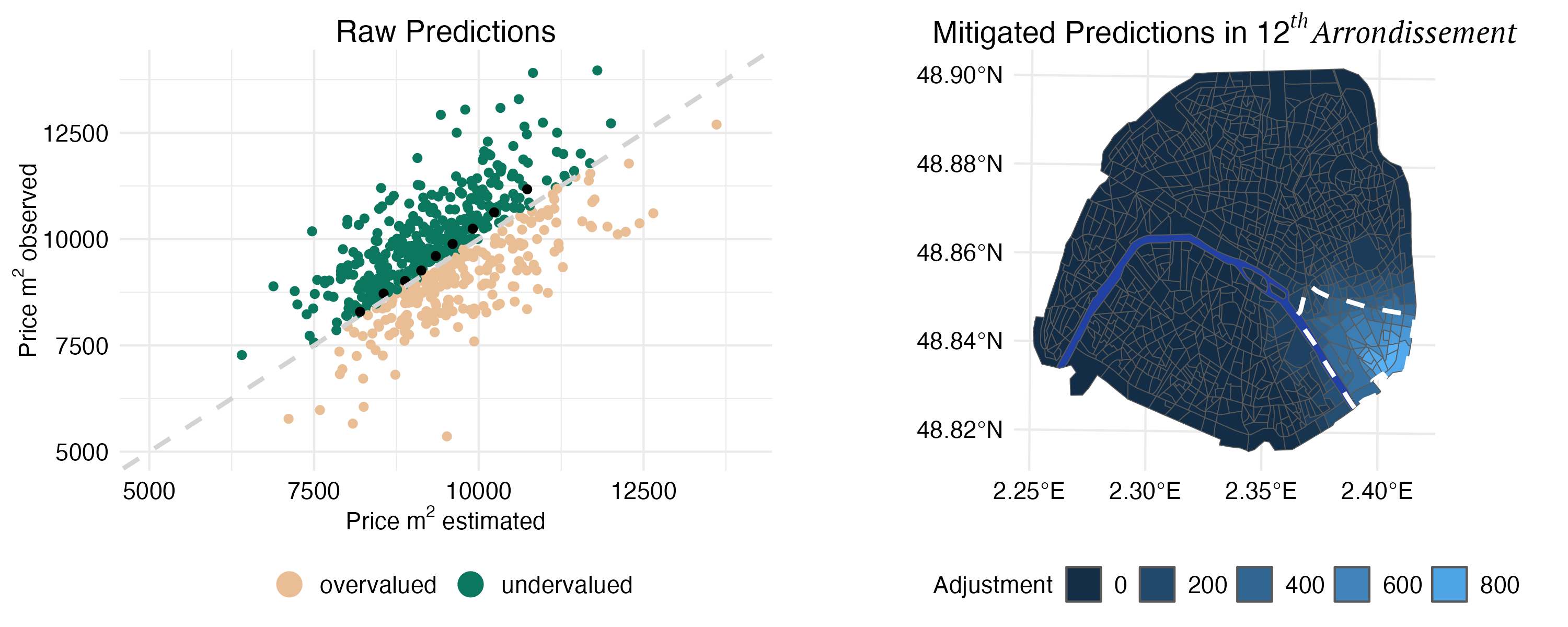}
    \caption{Predictions and mitigation changes when considering the 12$^{th}$ \emph{arrondissement} (boundaries shown in dashed lines).}
    \label{fig:mitigated_data}
\end{figure}

After fairness calibration, Figure~\ref{fig:mitigated_data} serves as an example, illustrating how the mitigation procedure changes the distributions of scores using data from the 12th \emph{arrondissement}.\footnote{The source code can be found at \href{https://github.com/fer-agathe/parisian_real_estate/}{(https://github.com/fer-agathe/parisian\_real\_estate/)}.} Here, around 60\% of all observations are underestimated by the model; hence we would expect the mitigation procedure to lift prices slightly. This is visualized in the left pane, where we visualized the whole set of mitigated prizes. The mitigation procedure works as expected, when comparing the results to Figure \ref{fig:smoothed_diffs}, one can observe that the regions most associated with a model underestimation get corrected the most.

Note that, with an appropriate post-processing technique, it is also possible to mitigate model uncalibration or unfairness under the EO constraint in this multi-class classification framework.

\section{Conclusion}

In this study, we analyzed the impact of spatial aggregation on both model calibration and estimation unfairness, and how to mitigate such effects. We found that the level of aggregation has a significant impact on the conclusions drawn from the same data. For example, although some granular levels of analysis deviate significantly from the overall calibration and fairness level, more aggregated data presented fewer issues. This naturally poses a conundrum when choosing the level of analysis using spatial data available at a more granular level. To smooth out estimates and regroup areas of analysis when the underlying geospatial data is of varying sizes, we developed a graph-based neighborhood construction that also allows smoothing out estimates, similar to kernel-based methods. We then proposed a methodology to mitigate effects stemming from under- or overvaluations in certain geographical areas. The proposed methods are both computationally efficient and work on both small and large data alike.

\begin{acks}
Ewen Gallic acknowledges that the project leading to this publication has received funding from the French government under the ``France 2030'' investment plan managed by the French National Research Agency (reference: ANR-17-EURE-0020) and from Excellence Initiative of Aix-Marseille University -- A*MIDEX. Fran\c{c}ois Hu acknowledges that the project is funded by the Natural Sciences and Engineering Research Council of Canada (NSERC) Emerging Infectious Diseases Modelling Initiative (EIDM), awarded to the Mathematics for Public Health (MfPH) program.
\end{acks}

\bibliography{bibliography}

\appendix

\section{Supplementary Graphical Results}

In this section, we provide additional figures to enhance understanding of the main content, offering complementary insights.

\subsection{Density of Observations}

Since the number of observations in our data are not uniformly distributed across all available IRIS codes, significant differences arise in the variance of, for example, mean estimates. Whereas regions on the northern bank of the Seine River appear to have a more liquid market, other regions to the east and south have fewer sales reported. 

\begin{figure}[H]
    \vspace{.1in}
    \centering
    \includegraphics[width=.5\linewidth]{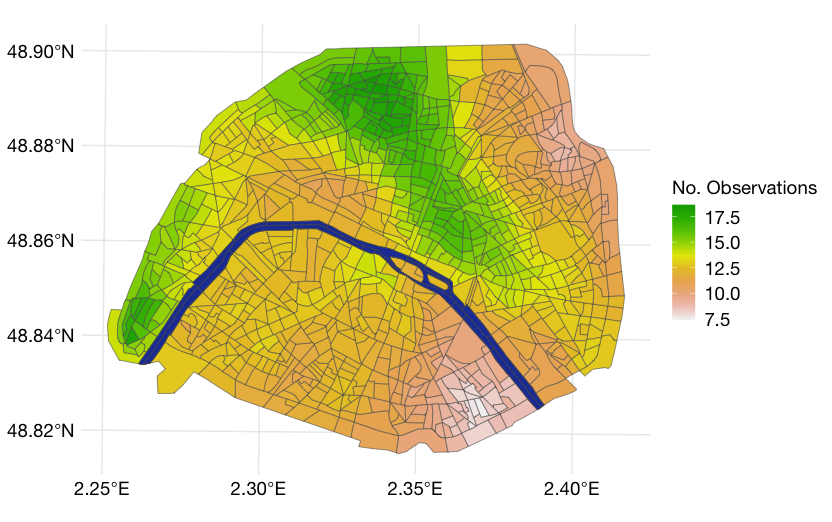}
    \caption{Number of available observations, smoothed using the neighborhood method.}
    \label{fig:smooth_obs_plot}
\end{figure}

\subsection{Wealth level}

As mentioned throughout the text, the wealth level per IRIS also corresponds roughly to the price per square meter. Many other indicators, such as the share of social housing also strongly correlated spatially with the average price per square meter of real estate. 

\begin{figure}[H]
    \vspace{.1in}
    \centering
    \includegraphics[width=.6\linewidth]{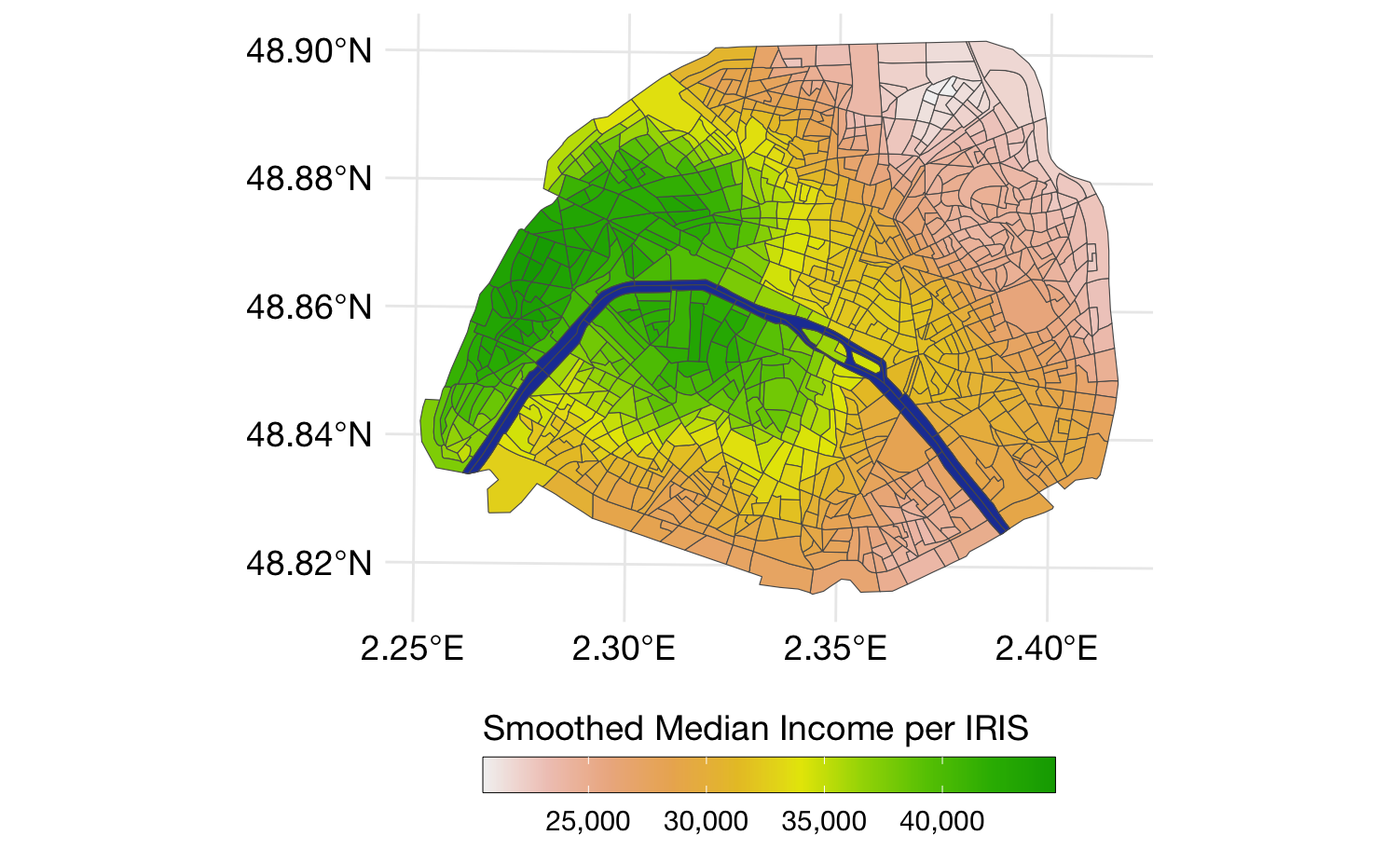}
    \caption{Estimated Income per IRIS region, smoothed using the neighborhood method. }
    \label{fig:smooth_med_inc}
\end{figure}

\subsection{Distribution of $m^2$ prices}

We opted to only include data from IRIS regions within the city of Paris itself, as there seem to be significant differences between the mean and shape of the overall distributions of square meter prices. 

\begin{figure}[H]
    \vspace{.1in}
    \centering
    \includegraphics[width=.6\linewidth]{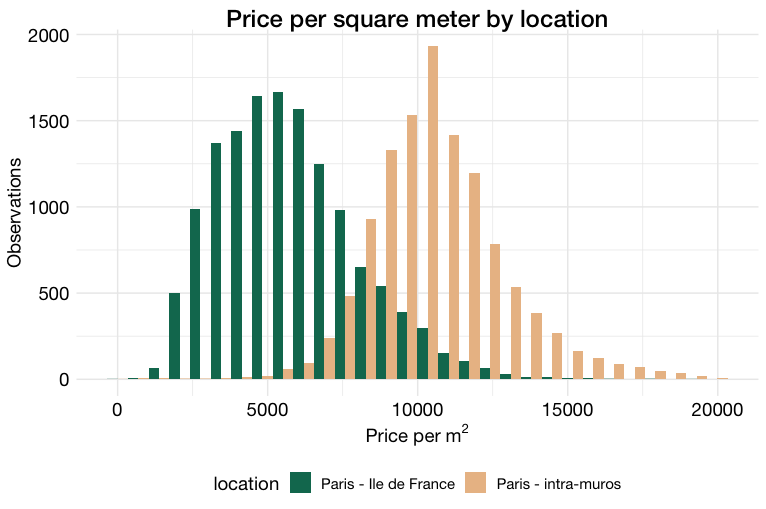}
    \caption{Price per square meter in different areas of our data. Paris \emph{intra-muros} refers to the 20 \emph{arrondissements} that constitute the core of the city, \emph{Ile de France} refers to the remaining metropolitan area. }
    \label{fig:plot_price_diff}
\end{figure}

\subsection{Calibration measurement}

\begin{figure}[H]
    \vspace{.1in}
    \centering
    \includegraphics[width=\linewidth]{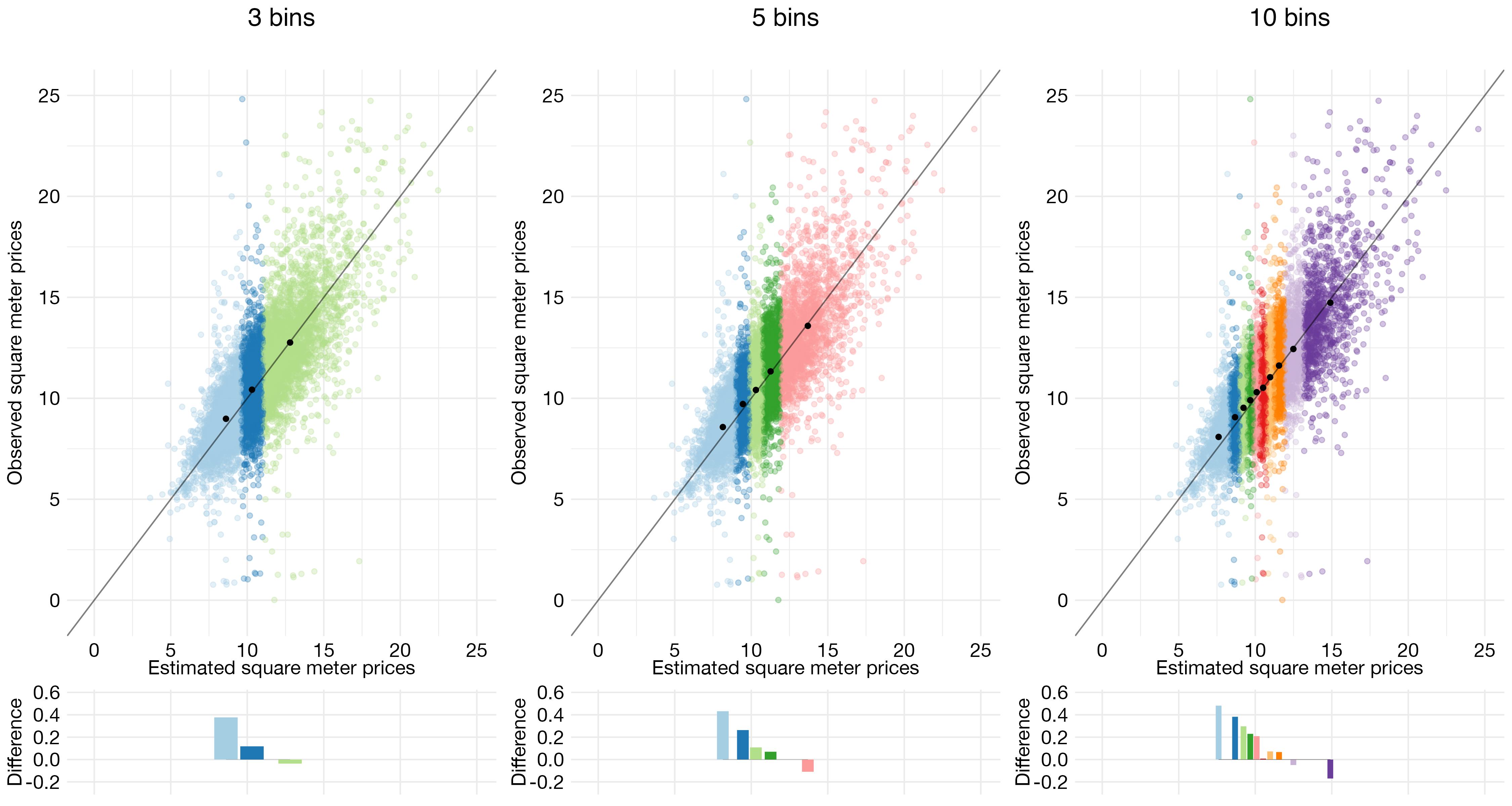}
    \caption{Calibration on the whole dataset estimated using different number of bins. Prices are in thousand Euros.}
    \label{fig:calibration_bins_whole}
\end{figure}



\appendix

\end{document}